\documentclass{article}
\usepackage{iclr2027_conference,times}

\usepackage{amsmath,amsfonts,bm}

\def\eqref#1{equation~\ref{#1}}

\def\1{\bm{1}}

\DeclareMathAlphabet{\mathsfit}{\encodingdefault}{\sfdefault}{m}{sl}
\SetMathAlphabet{\mathsfit}{bold}{\encodingdefault}{\sfdefault}{bx}{n}

\usepackage{hyperref}
\usepackage{url}
\usepackage{booktabs}
\usepackage{amsmath}
\usepackage{amssymb}
\usepackage{graphicx}
\usepackage{xcolor}
\usepackage{xspace}
\usepackage{enumitem}

\newtheorem{observation}{Observation}

\newcommand{\sys}{AGAR\xspace}

\title{\sys: A Reinforcement Learning Substrate\\for LLM Program Evolution}

\author{%
  \parbox{\textwidth}{\centering
    Haoran Li$^1$, Zengle Ge$^2$, Xiaomin Yuan$^2$, YUI LO$^2$, Haoxin Li$^3$, Songlin Zhou$^3$,
    Qianhui Liu$^4$, \\[2pt]
    Jiahua Ying$^2$, Yuanhang Liu$^3$, Mingju Chen$^5$, Annan Li$^2$,
    Jianmin Wu$^2$, Dawei Yin$^2$, Dou Shen$^2$ \\[6pt]
    {\small $^1$University of the Chinese Academy of Sciences \quad $^2$Baidu \quad $^3$Tsinghua University} \\[2pt]
    {\small $^4$Institute of Automation, Chinese Academy of Sciences \quad $^5$Shanghai University}
  }
}

\begin{document}
\AtBeginDocument{\lhead{}}

\maketitle

\begin{abstract}
Given a task and an evaluator, a language model can rewrite a candidate program
while a search loop decides which rewrites survive --- a practical route to
algorithm discovery. But that loop is governed by five constants set by
hand --- which parent to select, how hard to mutate, how to keep diversity, what
to remember, and a scalar score that never says which part of the program earned
it --- and reinforcement learning already has an estimator for each. The
obstacle is that program evolution is not usually written down as a decision
process. We
formalize it as a Markov decision process whose action is the \textbf{modular
prefix the model is conditioned on, not the program it emits}, so that credit
assignment, value estimation, adaptive exploration and experience memory each
attach to a distinct component. \sys (\emph{Algorithm Generation As RL})
provides the substrate this licenses: any estimator can be replaced or switched
off without touching the controller, making the transfer auditable one mechanism
at a time, with no gradient training of the backend model. Across 19 tasks, two
backends and three seeds under one harness, \sys improves on the stronger of two
published baselines on most tasks, with the gains concentrated on the
competitive-programming family.
The formalization also yields a checkable
reading of prior work: these systems are implicitly zero-discount, not by choice
but because fitness is exogenous to an individual rather than a return over
successors, leaving a discount factor nothing to act on.
\end{abstract}

\section{Introduction}
\label{sec:intro}

A long-horizon algorithm optimization task --- find a program that scores well on a
task-specific evaluator, given a budget of iterations and no gradient of the
score with respect to the program text --- is a sequential decision problem. It
is usually not \emph{written down} as one. It is written down as an evolutionary
loop: maintain a population of programs, ask a language model to mutate or
recombine them, evaluate the offspring, keep what scores well, repeat
\citep{funsearch2024,alphaevolve2025,shinkaevolve2025,openevolve2025}. The
results are real: improved bounds on open combinatorial problems, faster kernels,
better heuristics.

\paragraph{The thesis.} We take the position that such a task \emph{can be
modelled as an RL problem}, and that doing so is not a re-labelling exercise: the
modelling choices determine which machinery is applicable, and the machinery then
determines the system. Concretely, we formalize the loop as an MDP whose action is
the \emph{modular prefix} the model is conditioned on --- not the program it emits
--- and we show that under that choice each load-bearing idea of modern RL
\citep{sutton2018reinforcement} attaches to a distinct component of the loop.
$M_1$ (credit assignment), $M_2$ (value estimation), $M_3$ (adaptive exploration)
and $M_4$ (experience memory) are the places the modelling says the estimators go.
We are deliberate about the strength of that claim: the formalization forces which
\emph{kinds} of estimator the objective requires and whether each is an action
factor or retrieved context, but it does not force that there are four of them,
and the boundaries we draw between $M_1$, $M_2$ and $M_3$ are a factorization we
chose rather than one the mathematics uniquely picks out. What we contribute is a
\emph{general evolutionary substrate} that carries this decomposition: an interface
in which any one of the estimators can be replaced or switched off without touching
the controller, so that the transfer is auditable one mechanism at a time.

\paragraph{The gap and the obstacle.} What the LLM contributes today is the
mutation operator. Everything that decides \emph{where to mutate, how hard, and
what to remember} is a hand-set constant: five decisions, each with an estimator
RL already has. The obstacle is not reward density --- RL routinely operates under
terminal-only reward, and LLM-driven program evolution does receive intermediate
signal which the mutation operator consumes. The real difference is structural. In
RL the value of a state is defined \emph{recursively} over its successors, so
terminal credit is propagable in principle. In evolution, fitness is an
\emph{exogenous property of an individual}; no Bellman equation relates a parent's
fitness to its children's, so there is no recursion to propagate along at all. We
make this precise in \S\ref{sec:the-line}. The formalization below makes the
recursion \emph{definable} over the search without claiming to make the backup
\emph{empirical}: the value it recurses on is estimated in text space rather than
learned from observed returns, and that boundary stays exactly where it is.

\paragraph{A checkable reading of prior work.} Because the action is a prefix
conditioning a frozen black-box model, the objective is
$J(\pi) = \mathbb{E}[\sum_t \gamma^t r_t] + \lambda H(\pi)$.
Setting $\gamma{=}0$ recovers the selection rule every system in this literature uses
(Observation~\ref{obs:gamma0}): immediate fitness plus a population entropy term.
This is not a metaphor but a degeneracy — and not a criticism, since without a
recursion a discount factor has nothing to act on. Making the recursion definable
is what turns $\gamma$ into a tunable knob.

\paragraph{A substrate, not a system.} The substrate (\S\ref{sec:substrate})
carries the transfer: interchangeable modules, interchangeable execution backends,
native asynchrony. Its engineering problems are those of asynchronous RL
post-training stacks \citep{agentlightning2025,slime2025,miles2025,sao2026}; the
optimand differs (text, not weights) and there is no gradient.

\paragraph{Contributions.}
\begin{enumerate}[leftmargin=1.4em,itemsep=0pt,topsep=2pt]
\item \textbf{A modelling of algorithm optimization as RL and the substrate it
licenses} (\S\ref{sec:formalization}). Action-as-prefix makes five RL tools
expressible in text space, each replacing a named fixed rule in a specific module.
The $\gamma{=}0$ degeneracy falls out of the same formalization. The contribution
is an \emph{interface} auditable one mechanism at a time, not a uniqueness result.
\item \textbf{Modular TextGrad: policy improvement in text space}
(\S\ref{sec:textgrad}). Each module's persistent \emph{guidelines} --- the
text-space analogue of policy weights --- are optimized without gradient training.
We report a measurement of the operator: a local-edit and validation-revert design
that replaces the original rewrite operator (\S\ref{sec:tg-measurement}).
\item \textbf{Asynchronous execution and the credit-assignment problem it creates}
(\S\ref{sec:async}, \S\ref{sec:staleness}). The streaming scheduler creates
staleness: credit is assigned against a population that no longer exists. We
characterize $\tau$ empirically (naive $\tau\!\le\!W{-}1$ is refuted in practice)
and relate it to asynchronous RL post-training \citep{sao2026}.
\end{enumerate}

\section{Background: Two Independent Lineages}
\label{sec:background}

Evolutionary computation searches by maintaining a population, perturbing it, and
keeping what scores well. Genetic programming applied this to program spaces
decades ago \citep{koza1992gp}; the LLM era replaced the syntax-tree mutation
operator with a language model, which is what made the approach competitive on
real algorithm-design problems \citep{funsearch2024,alphaevolve2025}. Later
systems differ in how they manage the population --- ShinkaEvolve adds a bandit
over LLM backends and novelty sampling \citep{shinkaevolve2025}; SMC-based
variants replace generational selection with sequential Monte Carlo resampling
\citep{smcevolve2026}; EvoX makes the search strategy itself the object of
evolution \citep{evox2026} --- but they share the \emph{structure} of the signal.
Fitness attaches to a completed individual, not to a step of a trajectory, and
intermediate information informs the \emph{generation} of the next candidate
without ever accumulating into a return. The search state --- which lineages are
productive, which regions are exhausted --- is therefore carried implicitly in the
population rather than estimated.

Reinforcement learning developed a different toolkit because it assumed a
different structure, a trajectory over which value is recursively defined: credit
assignment propagates outcome information backwards along it \citep{sutton1988td};
value estimation predicts return before it is observed \citep{mnih2015dqn};
adaptive exploration makes the explore/exploit trade-off a decision rather than a
constant \citep{auer2002ucb,garivier2011ducb}; entropy regularization turns
diversity into an objective term \citep{haarnoja2018sac}; and experience replay
makes the run's history a learnable resource \citep{lin1992replay,schaul2016per}.
Our five transferred tools are these five, and \S\ref{sec:transfer} states what
each had to become.

\subsection{What actually separates them}
\label{sec:the-line}

It is tempting to say that RL has dense reward, evolution has sparse reward, and
that this is the obstacle. That is inaccurate in both directions: RL routinely
operates under terminal-only reward, and LLM-driven program evolution does receive
intermediate signal which the mutation operator consumes. The real difference is
structural. In RL, the value of a state is defined recursively in terms of its
successors, and terminal credit is in principle propagable along that recursion via
returns or TD($\lambda$) --- sparse reward makes this \emph{statistically} hard,
but the problem stays well-posed. In evolution, fitness is an \emph{exogenous
property of an individual}; a child is not a successor state reached by acting in
the parent's state but a new object emitted by a stochastic operator, and no
Bellman equation relates a parent's fitness to its children's. There is therefore
nothing to back up \emph{along}, which is why evolutionary search assigns credit
by \emph{survival} rather than by attribution.

The formalization of \S\ref{sec:formalization} supplies the missing
\emph{structure}: by defining the action as a prefix and the transition as one
generation step, it makes a Bellman recursion well-posed over the search. It does
not supply the missing \emph{signal}: our $V_{M_2}$ is an estimate produced by a
text-space heuristic, not a value learned by backing up observed returns. What
\emph{is} live is the forward half --- the estimate computed at dispatch ---
through two channels active in every reported run: the $M_3$ selection rule
$f(p) + \gamma V(p)$ and the $M_2$ lineage ranking. So:
\begin{center}
\emph{We make the recursion definable. We do not make the backup empirical.}
\end{center}
This also explains Observation~\ref{obs:gamma0}: prior systems are $\gamma{=}0$
not because their designers chose greediness, but because without a recursion a
discount factor has no successor value to act on.

\subsection{Positioning}

The recurring distinction between \sys and the closest prior systems is that a
rule which is fixed elsewhere becomes a learnable, replaceable component here
(Appendix~\ref{app:related} gives a row-by-row comparison). A parallel line of
work builds substrates for applying RL to agents at scale
\citep{agentlightning2025,slime2025,miles2025}, together with
asynchronous post-training systems such as SAO \citep{sao2026}. These are our
closest methodological relatives: they share our engineering problems ---
decompose the loop into independently swappable components, make rollouts
asynchronous, bound the staleness that asynchrony introduces --- but in every one
of them the learnable object is a set of model weights updated by a
policy-gradient estimator, whereas here the model is frozen and the learnable
object is the \emph{text} that conditions it.

\section{Formalization and the Transfer}
\label{sec:formalization}

\subsection{LLM-driven program evolution as an MDP}
\label{sec:mdp}

\begin{figure}[t]
\centering
\includegraphics[width=0.88\textwidth]{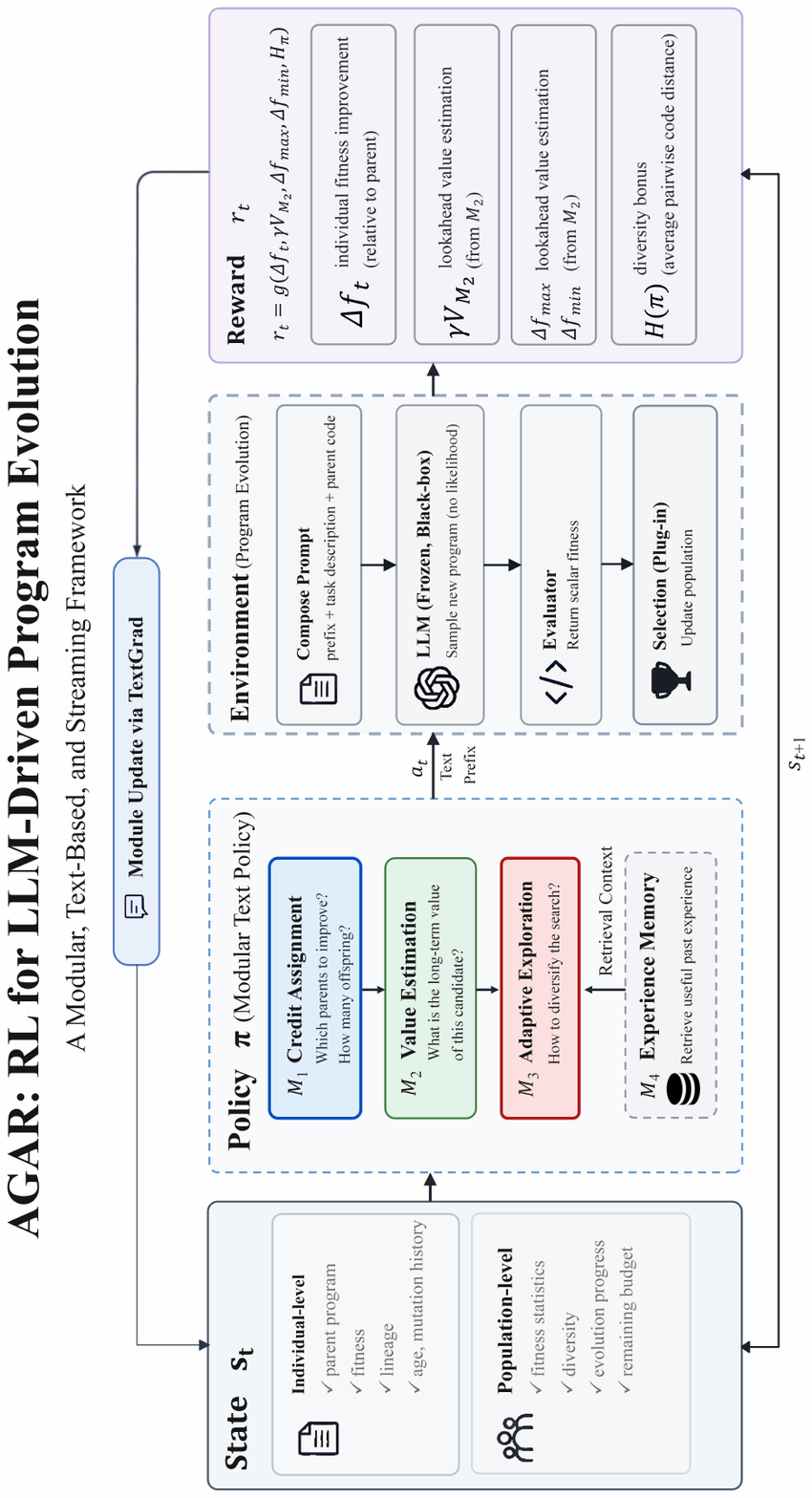}
\caption{The transfer as a whole. Left to right: the mixed-granularity state
$s_t$; the policy $\pi$, whose action $a_t$ is the \emph{modular text prefix}
$M_1$--$M_3$ produce, with $M_4$ supplying retrieved context rather than an
action factor; the environment, in which the frozen black-box LLM sees only
that prefix plus the task description and parent code, and a pluggable selection
operator folds the result back into the population; and the reward terms of
Eq.~\eqref{eq:reward}. The outer arrows are the two loops that make this RL
rather than a re-labelling: the TextGrad update over the module guidelines
$\theta_i$ (top), and the $s_t \to s_{t+1}$ transition the search itself
performs (bottom).}
\label{fig:framework}
\end{figure}

We model one evolutionary step as a transition in an MDP
$(\mathcal{S},\mathcal{A},P,R,\gamma)$. Figure~\ref{fig:framework} shows the
whole construction at once; the subsections below take its parts in turn.

\textbf{State $s_t$} is mixed-granularity by necessity: an individual-level part
(the parent program, its fitness, its lineage, its age, its mutation history) and a
population-level part (fitness statistics, a diversity measure, budget progress and
remainder). Both are required --- credit assignment and value estimation are
questions about an individual, exploration and entropy about a population --- and
neither subsumes the other: a purely individual state cannot express ``the
population has converged'', and a purely population-level state cannot express
``this lineage has stalled''.

\textbf{Action $a_t$ is a modular prefix, not a program.} This is the load-bearing
choice of the formalization. It is tempting to call the generated program the
action, but the program is sampled from a frozen black-box LLM: it is a
\emph{product of the transition}, not something the search policy emits. What the
policy emits is the text that conditions that sampling, so we define
\begin{equation}
a_t \;=\; \mathrm{Concat}\big(m_1(s_t),\, m_2(s_t),\, m_3(s_t)\big),
\label{eq:action}
\end{equation}
the concatenation of the credit-assignment, value-estimation and exploration
modules' natural-language outputs. $M_4$ (experience memory) does not enter the
prefix; it is retrieved context consumed by $M_1$--$M_3$. Two consequences follow,
and both are checks that the formalization does work rather than relabel: $\pi$ is
a text-emitting object improvable without touching model weights
(\S\ref{sec:textgrad}), and the module boundary is fixed by which factor of $a_t$ a
component writes rather than by taste.

\textbf{Transition $P$} has three stages: the execution LLM samples a program
conditioned on $a_t$ and the parent; an evaluator returns a scalar fitness; a
selection operator folds the offspring into the population. The first stage is a
frozen black box with no accessible likelihood, which forces a model-free treatment.
The third is \emph{pluggable}, which is where value-guided selection enters
(Table~\ref{tab:transfer}).

\textbf{Reward $R$} has individual and population terms:
\begin{equation}
\hat r_t \;=\;
\underbrace{\Delta f_t \;+\; \gamma\, V_{M_2}(\text{child})}_{R_{\text{individual}}}
\;+\;
\underbrace{w_{\max}\,\Delta f_{\max} \;+\; w_{\text{mean}}\,\Delta f_{\text{mean}}}_{R_{\text{population}}}
\;+\;
\underbrace{\lambda\,H(\pi)}_{\text{entropy}} .
\label{eq:reward}
\end{equation}
$\Delta f_t$ is the fitness improvement over the parent. The $\gamma V_{M_2}$ term is where
foresight enters: an estimate of what this child is worth as an \emph{ancestor}, filled
in retrospectively once its subtree exists (\S\ref{sec:substrate}). $\gamma$ is a fixed
scalar applied only inside Eq.~\eqref{eq:reward} --- never passed to the LLM as text,
and independent of the $\texttt{horizon}$ parameter that controls how much lineage the
modules see (see Appendix~\ref{app:reward} for defaults).

\subsection{The objective, and the \texorpdfstring{$\gamma{=}0$}{gamma=0} degeneracy}
\label{sec:objective}

\begin{equation}
\max_\pi \;\; J(\pi) \;=\; \mathbb{E}_{\tau\sim\pi}\Big[\textstyle\sum_t \gamma^t r_t\Big]
\;+\; \lambda H(\pi).
\label{eq:objective}
\end{equation}
We instantiate $H(\pi)$ as mean pairwise code distance within the population,
with a behaviour-class Shannon entropy as a secondary indicator. The choice is
deliberate: it costs zero additional LLM calls and is already monitored by $M_3$,
so the entropy term is measurable at every step rather than estimated. We add no
KL penalty; unlike RLHF there is no natural reference policy to stay close to.

\begin{observation}[Existing systems are the $\gamma{=}0$ special case]
\label{obs:gamma0}
Setting $\gamma{=}0$ in Eq.~\eqref{eq:reward} deletes the $V_{M_2}$ term and
collapses Eq.~\eqref{eq:objective} to
$\mathbb{E}[r_0] + \lambda H(\pi)$, i.e.\ a one-step objective in which an
individual is worth exactly its immediate fitness improvement plus its
contribution to population statistics. This is precisely the selection rule
used by every LLM program-evolution system we are aware of.
\end{observation}

Observation~\ref{obs:gamma0} is an algebraic identity, not a metaphor, and it is
the cleanest reading the formalization gives us of the prior art: $\gamma$ is a
knob whose zero setting reproduces a one-step objective, because a discount
factor with no successor value to act on has nothing else to be. We state it as
an identity over Eq.~\eqref{eq:reward} and leave the empirical consequences of
setting $\gamma > 0$ to future work.

We read this as a consequence of \S\ref{sec:the-line} rather than a shortcoming
of prior work. A discount factor needs a successor value; absent a recursion over
the search there is none, so $\gamma{=}0$ is the only setting available. What
\S\ref{sec:mdp} contributes is not a better choice of $\gamma$ but the structure
that gives $\gamma$ a referent at all --- with the honest counterpart that the
$V_{M_2}$ it multiplies is estimated in text space rather than bootstrapped from
observed returns.

\subsection{The transfer}
\label{sec:transfer}

Table~\ref{tab:transfer} in Appendix~\ref{app:transfer} is the centre of the
paper. Each row names an RL tool, the evolutionary fixed rule it displaces, the
module it becomes, and --- most importantly --- \emph{what had to change} given
the absence of an empirical backup (\S\ref{sec:the-line}).

We stress the direction of derivation. We did not choose four modules and then
find RL names for them; the four factors of $a_t$ in Eq.~\eqref{eq:action} plus
the retrieved context $M_4$ are what the transfer produces, and the fact that
$M_4$ is \emph{not} a factor of the action --- it is context --- is a
consequence of the formalization rather than an implementation detail.

\paragraph{Instantiations.}
A choice of implementation for each of $M_1$--$M_4$ defines a point in a
configuration space (\S\ref{sec:extensibility}). Throughout the paper we report
two such points, \textsf{Base} and \textsf{Combo}, differing in the $M_2$
estimator and the $M_3$ decision rule (statistical vs.\ LLM-based; rule-based
vs.\ discounted-UCB). They are sampled points, not competing systems; we do not
argue for one over the other.

\subsection{Critic independence, and what it forces}
\label{sec:critic}

Each module is improved against its \emph{own} critic, which judges only whether
that module's output was correct on its own terms --- was the attribution
consistent with the observed lineage? did the predicted ranking match the realized
ordering? --- and never attempts to apportion final fitness across modules. The
reason is structural: all four outputs are concatenated into one prefix
(Eq.~\ref{eq:action}) and consumed jointly by a black-box LLM, so their
contributions to the resulting program are not separable from the inside. This has
a design consequence: which transferred tool carries the gain can only be answered
by external leave-one-out ablation, not from inside the system. We do not report
that decomposition: at $n{=}3$ the level-to-level differences of
Table~\ref{tab:ladder} already sit inside the noise. See
Appendix~\ref{app:modules} for the full module data contracts.

\section{Modular TextGrad: Policy Improvement in Text Space}
\label{sec:textgrad}

Section~\ref{sec:formalization} defines what we optimize. This section defines
how: the method is TextGrad \citep{textgrad2024} applied to a target its original
formulation already fits --- an objective with no analytic gradient, computed by an
external system, over a variable that is itself text. The frozen backend model
writes code; each module's \emph{persistent guidelines} $\theta_i$ are the
variable being optimized; the loss is $\hat r$ of \S\ref{sec:formalization}; and
the ``gradient'' is the rollout context shown to the LLM critic. Appendix~\ref{app:textgrad}
gives the full operator definitions and the RL-to-text correspondence table.

Two features of our instance are unlike the tasks TextGrad was demonstrated on.
First, the variable is \emph{modular}: the policy is a concatenation of per-module
guidelines, so the gradient must be taken per module and a critique that conflates
them is unusable. Second, the loss is measured over a \emph{period} of noisy
evolutionary steps rather than a single forward pass, so the update rests on a
small-sample signal and must decide what to do when that signal is inconclusive.
The second feature is what \S\ref{sec:tg-measurement} is about.

\subsection{Operators}
\label{sec:tg-operators}

For each module $i$ we define a per-module textual gradient operator $G_i$ and a
shared update operator $U$ (full definitions in Appendix~\ref{app:textgrad}).
The object $\theta_i$ is the module's \emph{persistent guidelines}: a block of
text that survives across evolutionary steps and conditions every subsequent
invocation of that module --- not the module's per-step output, which is an action,
not a policy parameter. Optimizing the output instead of the guidelines would fail
to persist and therefore cannot constitute policy improvement.

\subsection{What we measured, and why the operator was replaced}
\label{sec:tg-measurement}

The rewrite operator --- critique in, new guidelines out, unconditionally --- is
the direct transposition of TextGrad to this setting. Two measurements over 206
runs (9{,}844 update pairs) forced its replacement. \textbf{(i) The variable was
resampled, not improved}: the guideline block hit its 2000-character budget at the
\emph{first} update of every run and stayed there; textual similarity between
adjacent versions was 0.18, between first and last 2--6\%, without exception.
\textbf{(ii) Nothing closed the loop}: a rewrite that made things worse persisted
until the next rewrite replaced it, so there is no mechanism by which $\theta_i$
at step 100 should beat $\theta_i$ at step 1.

\textbf{The repaired operator} replaces $U$ with a local edit plus deferred
validation. The policy is an explicit numbered rule list; $G_i$ returns at most
one \textsf{ADD} and one \textsf{EDIT}/\textsf{REMOVE} per period
(\textsf{NONE} is an expected answer); validation is deferred $\texttt{verdict\_lag}$
periods because a new rule's children are mostly leaves at install time, making
immediate scoring equivalent to $\gamma{=}0$. Cost falls from $8$ to $2$ LLM
calls per update (only $M_1$ and $M_4$ are edited; the other two were writing
guidelines no code path read). The budget is bounded by construction, and the
operator became legible: discrete rules with provenance can be reported at run end,
which the rewrite operator made impossible. Full measurement detail is in
Appendix~\ref{app:textgrad}.

\section{Substrate: Extensibility and Asynchronous Execution}
\label{sec:substrate}

A modelling choice only becomes a system if the interfaces it implies are real.
This section describes the two properties of the substrate the transfer actually
depends on --- replaceable components and native asynchrony --- and one problem
the second creates for the first. We develop the asynchronous part against the
closest engineering precedent, the asynchronous RL post-training systems surveyed
in \S\ref{sec:background}.

\begin{figure}[t]
\centering
\includegraphics[width=0.88\textwidth]{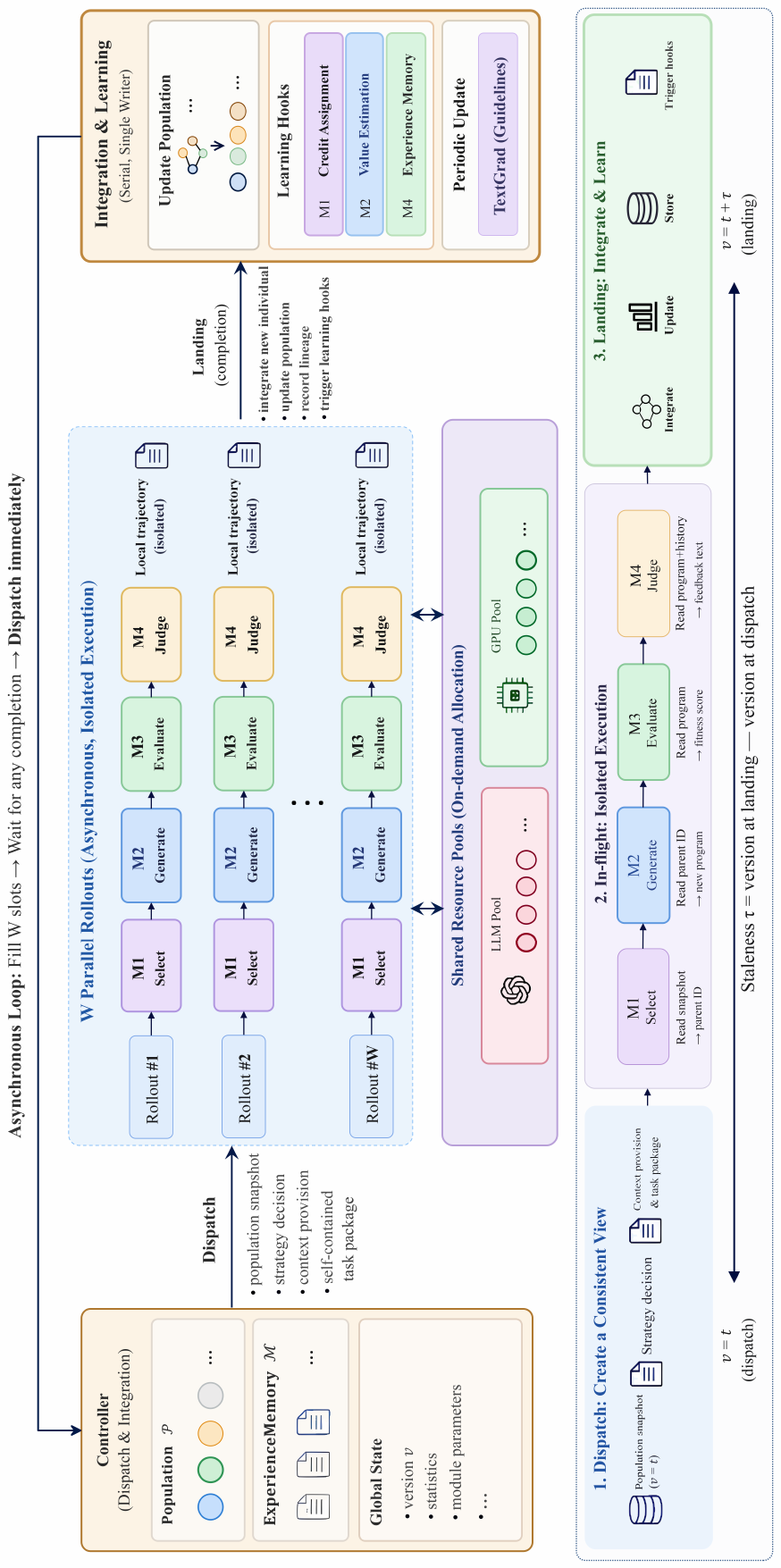}
\caption{The substrate. Top: the streaming loop --- fill $W$ slots, wait for
\emph{any} completion, dispatch a replacement immediately --- with the controller
as sole writer of the population and the $W$ in-flight rollouts reading a
dispatch-time snapshot and writing only their own trajectory. The four pipeline
stages (\textsc{select}, \textsc{generate}, \textsc{evaluate}, \textsc{judge})
are the substrate's; the replaceable modules sit in the \textsc{select} slot.
Shared resource pools are acquired per use and returned. Bottom: the same loop
in the three moments used throughout this paper, with $\tau$ defined as the
population-version difference between dispatch and landing.}
\label{fig:substrate}
\end{figure}

\subsection{Extensibility, stated so that it can be checked}
\label{sec:extensibility}

``Extensible'' is easy to assert, and every substrate paper asserts it. We give
three measurable forms. (a)~\emph{Configuration space}: the choice of
implementation for $M_1\times M_2\times M_3\times M_4$ spans 180 configurations,
of which we have evaluated 14; \textsf{Base} and \textsf{Combo} are two of those
points. (b)~\emph{Integration cost}: replacing $M_3$'s LLM explore/exploit call
with the discounted-UCB bandit of \citet{garivier2011ducb} added 236 lines in one
new module file and changed 0 lines in the controller or the module protocol ---
and \emph{deleted} 26 lines of hand-written budget and stagnation heuristic that
the discounting subsumes. That deletion is the part we would point to: the claim
is not that new mechanisms are cheap to add, but that adding a principled one can
remove a hand-tuned constant.

(c)~\emph{Interchangeable backends}, and a non-obvious cost. Backends register
through a single entry point and can be swapped between a thread pool and a
distributed runtime. During development this produced a failure worth reporting:
under the distributed backend, per-task serialization gives each rollout an
isolated module instance for free, while under the thread-pool backend a shared
evaluator instance guarded by an environment lock caused rollouts to be
\textbf{silently serialized} --- no error, no warning, and worse the longer each
evaluation takes, i.e.\ exactly on the workloads where concurrency matters most.
The fix was explicit per-task isolation of module state. The general lesson is
that \emph{extensibility is not freedom; it is an obligation on the interface to
state its isolation semantics}. An interface permitting a backend swap without
specifying whether state is shared has not made the system extensible, only
replaceable. Appendix~\ref{app:extensibility} gives the integration diff in full.

\subsection{Asynchronous execution}
\label{sec:async}

The scheduler is streaming rather than generational: it fills $W$ in-flight
rollouts, waits for the first to complete, folds the result into the
population, and immediately dispatches a replacement. There is no batch
barrier, so a single slow evaluation delays one slot rather than the whole
generation --- which matters here because per-rollout times across the tasks we study
span $3.7\times$ (median $269$\,s on the fastest task against $1007$\,s on the
slowest), so a straggler holds a slot for several times the median. As an engineering decision this needs no
justification: it is what every production RL stack does, and the reasons are
the same.

The interesting part is what the asynchrony does to the \emph{learning rule},
and it is here that the evolutionary setting diverges from the RL one.

\paragraph{Comparison with asynchronous RL post-training.}
The asynchrony forces single-rollout consumption, which is the native form for
evolutionary rollouts (each offspring is scored against the population, not against
siblings) rather than a concession. SAO \citep{sao2026} reaches the same design
from the post-training side; Appendix~\ref{app:extensibility} details the two
further points of contact that look like correspondences but are not.

\subsection{Staleness: what asynchrony does to credit assignment}
\label{sec:staleness}

The dispatch path samples a parent, computes the $M_2$ estimate, and runs the
$M_3$ decision \emph{at dispatch time}; the resulting fitness lands some number
of population updates later. Call that gap $\tau$, so $\tau \equiv 0$ is the
synchronous case. $\tau$ has \emph{no} a priori upper bound: because the
scheduler is streaming, a slow rollout can outlive an arbitrary number of faster
ones, and $\tau$ is bounded only by the tail of the evaluation-time distribution.
Bounding it at $W{-}1$ would require the generational synchronisation we rejected
for throughput. We therefore characterise $\tau$ empirically rather than
bounding it; the measured distribution is in Appendix~\ref{app:staleness}, where
at $W{=}8$ the maximum is $22$ against a naive bound of $7$.

The consequence connects to Table~\ref{tab:transfer}: credit is assigned against
a population that no longer exists. $M_2$'s baseline statistics are $\tau$ updates
old; $M_3$'s selection rule consumes a value estimate computed against a
superseded population; $M_4$'s novelty distances are computed against a stale
population; and the Meta layer consumes a rollout buffer produced under
superseded guidelines --- which is precisely the sense in which
\S\ref{sec:textgrad}'s operator is off-policy. The standard correction does not
transfer: SAO's clipped importance ratio needs the two policies to be comparable
\emph{per token}, and our ``policy'' is a block of natural language that an update
may replace wholesale. A clipping-style correction would require first defining a
distance between two texts and arguing that it bounds a likelihood ratio, and we
have no such argument. We report the delay instead of correcting it.

\section{Experiments}
\label{sec:experiments}

We ask two questions, both of which the formalization makes answerable:
(i)~does the transfer help at all, under one harness and against published
systems? (ii)~how much of that is the accumulated substrate? The third
component of the paper's evidence is Observation~\ref{obs:gamma0} itself, which
is an algebraic identity over Eq.~\eqref{eq:reward} and therefore requires no
experiment; we state it here as a fact about the construction and leave the
empirical consequences of the discount to future work.

\subsection{Setup}
\label{sec:exp-setup}

\textbf{Tasks.} 19 tasks in three families: 10 competitive-programming
heuristic-optimization problems (C++), 3 mathematical construction problems, and
6 systems/ML algorithm-discovery problems (Python). Median per-rollout time
differs across tasks by $3.7\times$ ($269$--$1007$\,s) and score scales differ
entirely, so all aggregation is on per-task normalized scores.

\textbf{Models, seeds and baselines.} 2 backend models (DeepSeek-V4-Flash, GLM-5.2)
$\times$ 3 seeds; we treat $n{=}3$ as supporting qualitative conclusions only.
Baselines are OpenEvolve and ShinkaEvolve, run under their own harnesses with task
specifications ported by us, plus \textsf{vanilla} --- our substrate with every
transferred module disabled. \textsf{vanilla} is the scientifically important
baseline; the external systems are the comparability check.

\subsection{Quality at matched budget}
\label{sec:exp-main}

\begin{table}[t]
\caption{Best-at-budget, normalized mean$\,\pm\,$std across all 19 tasks (3 seeds per system).
Each task score is divided by the per-task maximum over all four systems before averaging,
so the best system on each task scores 1.0 and the number is comparable across tasks.
Per-task raw scores are in Appendix~\ref{app:full-results}. $\diamond$ two of the 19 tasks
have fewer than 3 seeds completed for \sys (\textsf{ahc015}, \textsf{ahc027}).}
\label{tab:main}
\begin{center}\small
\begin{tabular}{@{}lcccc@{}}
\toprule
\textbf{Backend} & OE & SE & \sys\,\textsf{Base} & \sys\,\textsf{Combo} \\
\midrule
DeepSeek-V4-Flash & $0.855{\pm}0.193$ & $0.919{\pm}0.138$ & $0.949{\pm}0.107$ & $\mathbf{0.990{\pm}0.024}^\diamond$ \\
GLM-5.2           & $0.897{\pm}0.131$ & $0.929{\pm}0.093$ & $\mathbf{0.979{\pm}0.045}$ & $0.972{\pm}0.052$ \\
\bottomrule
\end{tabular}
\end{center}
\end{table}

Both instantiations of \sys rank above both baselines on DeepSeek; Base leads on GLM with Combo
close behind. The standard deviation of \sys is noticeably lower than that of the baselines,
partly because five tasks are near-ceiling for all systems (\textsf{circle}, \textsf{cloudcast},
\textsf{erdos}, \textsf{first\_ac}, \textsf{prism}): on those tasks every normalized score is
close to 1.0, which suppresses variance. The meaningful signal is on the remaining 14 tasks,
and the ALE (competitive programming) family carries most of the gain.

\subsection{Does the transfer work, and how much is the substrate?}
\label{sec:exp-ablation}

The mechanisms this substrate carries fall into two groups that have to be
ablated separately, because each group is held fixed inside the other group's
comparison. The substrate layers ($L_0$--$L_3$) are active in both arms of any
module ablation and so cancel in its difference; conversely the transferred
modules are active at every substrate level. This section reports the substrate
ladder, which is the comparison we can make with complete data. The
module-level decomposition is a separate measurement, and
\S\ref{sec:discussion} explains why we do not report it.

\paragraph{The substrate ladder, accumulated.}
Each level adds exactly one thing to the one below it. $L_0$ is repeated
sampling --- rewrite the current best and nothing else; $L_1$ adds inspiration
samples and a natural-language critic, i.e.\ naive evolution; $L_2$ adds a
prioritized replay buffer (island pruning, purge, diversity slots); and $L_3$
adds sample-level credit assignment (a per-test-case weakness report). $L_2$
and $L_3$ are exactly the \textsf{Base} and \textsf{Combo} of
Table~\ref{tab:main}, so this ladder is also what separates those two
instantiations.

\begin{table}[h]
\caption{Substrate ladder (raw task score, mean over 3 seeds, every cell run to
the full 100 fitness evaluations). Backend: DeepSeek-V4-Flash.}
\label{tab:ladder}
\begin{center}\small
\begin{tabular}{@{}lrrrrr@{}}
\toprule
\textbf{Task} & $L_0$ & $L_1$ & $L_2$ & $L_3$ & $L_0\!\to\!L_3$ \\
\midrule
ahc008        & 0.5441 & 0.7380 & 0.5452 & 0.9041 & $+66.2\%$ \\
ahc015        & 0.9687 & 0.9835 & 0.9336 & 0.8284 & $-14.5\%$ \\
ahc016        & 0.6670 & 0.7316 & 0.9160 & 1.0188 & $+52.7\%$ \\
ahc024        & 0.8659 & 0.8819 & 0.8774 & 0.8636 & $-0.3\%$ \\
ahc039        & 0.8797 & 0.8978 & 0.9404 & 0.8843 & $+0.5\%$ \\
\midrule
\textbf{Mean} & \textbf{0.7851} & \textbf{0.8465} & \textbf{0.8425}
              & \textbf{0.8998} & $\mathbf{+14.6\%}$ \\
\bottomrule
\end{tabular}
\end{center}
\end{table}

This ladder must be read carefully, and we read it narrowly. End-to-end
($L_0 \to L_3$) the mean gains $+14.6\%$, and $L_0$ is a genuine floor: it is
the substrate running a bare rewrite loop with every accumulated mechanism
switched off. That is the comparison which shows the mechanisms are doing work
rather than the harness.

But \textbf{per task the ladder is not monotone}, and we do \emph{not} claim
that each added level helps. $L_1 \to L_2$ falls on three of five tasks and the
overall mean dips slightly ($0.8465 \to 0.8425$); ahc015 declines from $L_1$
through $L_3$ ($0.9835 \to 0.8284$) and ahc024 is flat. Two of the five tasks
carry essentially the whole mean gain (ahc008 $+66.2\%$, ahc016 $+52.7\%$). At
$n{=}3$ the level-to-level differences sit inside the noise, and the per-task
non-monotonicity is the direct evidence of that. What the table supports is
that the accumulated stack is worth $+14.6\%$ over the bare loop --- not an
attribution of that figure to any one level.

\section{Discussion}
\label{sec:discussion}

\paragraph{What the transfer does and does not establish.}
The formalization of \S\ref{sec:formalization} buys three things: a precise sense
in which the prior art is a special case (Observation~\ref{obs:gamma0}), a
factorization of the loop into modules that the objective constrains but does not
uniquely determine, and a vocabulary in which mechanisms from the RL literature can
be stated, implemented and compared inside an evolutionary loop. What it buys, in
one sentence, is that the Bellman recursion becomes \emph{definable} over
evolutionary search. What it does not buy is an \emph{empirical} backup: the value
the recursion consumes is estimated in text space rather than learned from observed
returns. The two data-backed limitations and the full unverified list appear in
Appendix~\ref{app:limitations}.

\paragraph{Future work.}
The most direct next step is wiring the TD correction into the loop, turning $M_2$
from a one-shot estimator into a learned one.
Beyond that: importance correction for staleness \citep{sao2026};
cross-model critics; leave-one-out module ablation at larger $n$;
standardizing the $V$ scale so $\gamma{>}0$ can be tested cleanly; and using
the 180-configuration space of \S\ref{sec:extensibility} to ask which regions of
mechanism space are worth exploring --- a question that only becomes askable once
mechanisms are interchangeable.

\bibliography{references}
\bibliographystyle{iclr2027_conference}

\appendix
\section{Statement on the Use of AI Assistants}
\label{app:ai-use}

LLMs were used in this work in two distinct roles, which we separate
deliberately.

\paragraph{As the object of study.}
The system described in this paper invokes large language models as its
mutation operator, as its module policies, and as the critics of
\S\ref{sec:textgrad}. These uses are the subject matter of the paper and are
documented in \S\ref{sec:formalization}--\S\ref{sec:substrate} and
Appendix~\ref{app:reproducibility}, including model identifiers, decoding
parameters and call accounting.

\paragraph{As a writing and coding aid.}
The authors used an LLM-based assistant for implementation and analysis
support: refactoring module code, writing the experiment submission and
result-aggregation scripts, computing the summary statistics reported in
\S\ref{sec:exp-ablation} from the raw run artifacts,
and revising LaTeX prose. It was also used to audit consistency between the
manuscript and the implementation, which is how several discrepancies between
the two were located. All
scientific claims, experimental designs, analyses and conclusions are the
authors'. The authors verified every reported number against the raw
experimental artifacts and take full responsibility for the content of the
paper, including any errors.

No AI system is listed as an author.

\section{Reproducibility Statement}
\label{app:reproducibility}

\S\ref{sec:exp-setup} gives the task suite, the two backend models and the
budget. Table~\ref{tab:main} reports the normalised mean across all 19 tasks per
condition (each task score divided by the per-task maximum, then averaged); per-task
raw scores and the full two-backend detail are in Appendix~\ref{app:full-results}.
Table~\ref{tab:ladder} reports the same for each substrate level. Seeds are fixed at $42$, $123$
and $456$ throughout and control population initialization and sampling order;
every condition is three seeds $\times$ $100$ fitness evaluations, with no
exceptions.

The metric is best-at-budget on each task's own objective, normalized per task
for aggregation only. The saturation criterion --- all four conditions within
$1\%$ --- was applied before ranks were inspected, and the tasks it excludes are
marked in Table~\ref{tab:main}. Decoding uses temperature $0.7$ and a
$393{,}216$-token cap; $\mathrm{top}\text{-}p$ is left at the serving default
rather than pinned, which we note because it is part of the sampling
distribution we do not control. Both backends are hosted models rather than
frozen artifacts, so exact reproduction of absolute scores may not be possible;
we regard the relative comparisons as the reproducible content.

\S\ref{sec:exp-ablation} states which conclusions are limited by sample size:
at $n{=}3$ the level-to-level differences of the substrate ladder sit inside the
noise, which is why we report the end-to-end figure and decline to attribute it
to any single level, and why no per-module leave-one-out decomposition is
reported. Baseline numbers are obtained by us running OpenEvolve and ShinkaEvolve under
our own evaluation protocol, with task specifications ported by us, rather than
copied from the cited papers. Those ports are part of the release, since the
baseline conditions are other people's systems and should be independently
auditable. Runs execute in a container image released with the code, each
allocated four CPU cores and $16$\,GiB of memory with no GPU; per-run LLM call
counts and token counts are logged. The
framework, all module implementations, every task specification, the exact
configuration file for each reported run, and the raw logs will be released
publicly upon publication. Each run directory in the release contains the
configuration actually used, so any reported cell traces to the settings that
produced it rather than to a described protocol.

\section{Ethics Statement}
\label{app:ethics}

This work studies automated algorithm discovery on benchmark optimization
problems: no human subjects, no personal data, no demographically sensitive
attributes. Every generated program is executed in a sandboxed evaluation
environment and none is deployed. Compute expenditure is the one cost we
consider material, and we quantify it rather than elide it.

\section{Additional Material}
\label{app:extra}

\subsection{Module interfaces}
\label{app:modules}

Each module is a class implementing the substrate's \texttt{Module} protocol,
invoked with two arguments: a read-only \texttt{context} (the experiment-level
world --- population snapshot, configuration, accessors) and a writable
\texttt{result} (the per-rollout trajectory). That split is what makes the
concurrency safe (\S\ref{sec:async}): a worker never writes the world, and the
controller never reads a worker's intermediate state.

The data contract below is taken from the implementations rather than
paraphrased, because it is what makes the module boundary checkable: a
component's identity is which field of $a_t$ it writes.

\begin{center}\small
\begin{tabular}{@{}lll@{}}
\toprule
\textbf{Module} & \textbf{Reads} & \textbf{Writes} \\
\midrule
$M_1$ credit
  & population accessor
  & \texttt{m1\_prefix} (str) \\
  & \texttt{rl\_evolve} metadata & \\
\midrule
$M_2$ value
  & population accessor
  & \texttt{m2\_ranking} (list) \\
  & \texttt{child\_stats}
  & \texttt{m2\_v\_scores} (dict) \\
  & lineage tree
  & \texttt{m2\_prefix} (str) \\
\midrule
$M_3$ exploration
  & population accessor
  & \texttt{result.selection} \\
  & \texttt{m2\_ranking}
  & \texttt{m3\_prefix} (str) \\
  & \texttt{m2\_v\_scores}
  & \texttt{m3\_decision} (str) \\
\midrule
$M_4$ memory
  & rollout history
  & retrieved context for $M_1$--$M_3$ \\
\bottomrule
\end{tabular}
\end{center}

Two structural facts are visible in this table and are worth stating because
they are the content of \S\ref{sec:formalization}'s claim rather than
implementation detail. First, $M_3$ is the only module that writes
\texttt{result.selection}: it is the sole Actor, while $M_2$ writes a ranking
and never a decision. Second, $M_4$ writes no prefix field --- it is retrieved
context consumed by the other three, which is why it carries no $\theta_i$ and
is not a factor of the action.

The modules execute strictly in sequence, $M_1 \to M_2 \to M_3$, then
generation, evaluation and judging. $M_3$ depends on $M_2$'s ranking; $M_1$ and
$M_2$ are mutually independent and could in principle run concurrently, but the
implementation does not do this and we do not claim it.

\subsection{Reward and entropy details}
\label{app:reward}

Defaults for Eq.~\eqref{eq:reward}: $\gamma = 0.95$ (with $\gamma = 0$ as the
degenerate arm of Observation~\ref{obs:gamma0}), $w_{\max} = 0.7$,
$w_{\text{mean}} = 0.3$, $\lambda = 0.1$.

$H(\pi)$ is instantiated as the mean pairwise cosine distance between the
feature vectors of population members, $1 - \langle v_i, v_j\rangle /
(\|v_i\|\|v_j\|)$, averaged over all pairs with both vectors present. Programs
without a feature vector are skipped rather than imputed. The quantity costs no
additional LLM calls and $M_3$ already monitors it for its diversity floor,
which is why the entropy term is measured at every step rather than estimated.

We report no sensitivity sweep over $\lambda$. The weights above were fixed
before the reported runs and not tuned against them; given that $n{=}3$ leaves
the level-to-level differences of Table~\ref{tab:ladder} inside the noise, a
sweep at this sample size would not be able to separate $\lambda$ settings
either, and we prefer to state that than to report a sweep that cannot resolve
what it appears to measure.

\subsection{Task suite}
\label{app:tasks}

Nineteen tasks in three families. The competitive-programming family is drawn
from AtCoder Heuristic Contest problems (\textsf{ahc008}--\textsf{ahc046}),
scored by each contest's own objective. The mathematical-construction family
(\textsf{erdos}, \textsf{first\_ac}, \textsf{second\_ac}) scores a constructed
object against a target bound. The systems/ML family (\textsf{circle},
\textsf{cloudcast}, \textsf{eplb}, \textsf{llm\_sql}, \textsf{prism},
\textsf{txn}) scores a discovered algorithm on a held-out workload.

Score scales are not comparable across tasks --- \textsf{prism} reports values
near $26$, \textsf{cloudcast} near $0.0016$, \textsf{txn} in the thousands ---
so every aggregate in this paper is computed on per-task normalized scores and
every table reports raw scores per task. Median per-rollout wall-clock time
ranges from $269$\,s (\textsf{eplb}) to $1007$\,s (\textsf{ahc016}), a
$3.7\times$ spread; this is the straggler profile that motivates streaming
dispatch (\S\ref{sec:async}).

Six tasks are classified \emph{saturated} --- \textsf{circle},
\textsf{cloudcast}, \textsf{erdos}, \textsf{first\_ac}, \textsf{prism} on both
backends, plus \textsf{ahc026}, which reaches the same ceiling by $200$
evaluations. The criterion is that all four conditions
fall within $1\%$, and it was applied before ranks were inspected. The
independent evidence that this classification is not post-hoc is
the behaviour of the substrate ladder, where the saturated tasks cluster
at a ceiling across every level.

\subsection{Staleness measurements}
\label{app:staleness}

Table~\ref{tab:staleness} gives the empirical distribution of $\tau$ under the
$W$-sweep. $\tau$ is computed at landing time as the
population-version at landing minus the version stamped at dispatch. At $W{=}1$
the scheduler is synchronous, so $\tau \equiv 0$ strictly --- this is a
calibration check on the instrumentation itself; every row with $W{>}1$ is
therefore a measurement, not an accounting artefact.

\begin{table}[h]
\caption{Distribution of $\tau$ (number of population updates between dispatch
and landing), pooled over 3 seeds per configuration. The last column is the
fraction of rollouts that exceed the naive bound $W{-}1$.}
\label{tab:staleness}
\begin{center}\small
\begin{tabular}{@{}llrrrrrr@{}}
\toprule
\textbf{Task} & $W$ & $n$ & \textbf{Mean} & \textbf{Median} & \textbf{p95} & \textbf{Max} & $>\;W{-}1$ \\
\midrule
ahc011 (C++) & 1 & 60  & 0.00 & 0  & 0  & 0  & 0.0\% \\
ahc011 (C++) & 2 & 330 & 0.90 & 1  & 2  & 5  & 13.0\% \\
ahc011 (C++) & 4 & 322 & 2.74 & 3  & 5  & 10 & 20.8\% \\
ahc011 (C++) & 8 & 333 & 6.05 & 6  & 12 & 22 & 18.6\% \\
\midrule
eplb (Python) & 1 & 250 & 0.00 & 0  & 0  & 0 & 0.0\% \\
eplb (Python) & 2 & 306 & 0.97 & 1  & 2  & 3 & 15.4\% \\
eplb (Python) & 4 & 90  & 2.46 & 2  & 5  & 6 & 21.1\% \\
\bottomrule
\end{tabular}
\end{center}
\end{table}

Two patterns are worth noting. First, the mean tracks $0.88\,(W{-}1)$ on both
tasks across the full sweep --- well below the naive $W{-}1$ scaling but clearly
non-zero, so staleness is real and cannot be treated as negligible. Second, the
tail is heavy and does not converge: at $W{=}8$ roughly one in five rollouts
exceeds $W{-}1$, and the maximum of $22$ is more than three times the naive
bound, a value that would require generational synchronisation to suppress.

Completion detection is by insertion-order scan with a $5$\,s polling timeout,
so rollouts finishing inside the same poll window fold into the population in
dispatch order rather than true completion order. Against a median rollout time
of several hundred seconds this biases the measured distribution negligibly, but
it is a measurement convention rather than an exact observable and we state it
as such.

The originally planned second column --- the relationship between $\tau$ and
$M_2$ estimation error --- \textbf{is not reported}. $\tau$ is written to the
run log but is not threaded into $M_2$'s statistics, so the per-rollout
correspondence between the two quantities is not available. Establishing it
would require plumbing the dispatch version number through to the value
estimator, which we have not done.

\subsection{TextGrad operator definitions}
\label{app:textgrad}

For each module $i$ the textual gradient operator is
$G_i:\; \mathcal{T}\times\mathcal{C}\times\mathbb{R} \rightarrow \mathcal{T}$
and the update operator is
$U:\; \mathcal{T}\times\mathcal{T} \rightarrow \mathcal{T}$,
where $\mathcal{T}$ is the space of natural-language strings and $\mathcal{C}$ is
the rollout context accumulated since the last update. One update cycle is
\begin{equation}
g_i^{(t)} = G_i\!\big(\theta_i^{(t)},\, c^{(t)},\, \hat r^{(t)}\big),
\qquad
\theta_i^{(t+1)} = U\!\big(\theta_i^{(t)},\, g_i^{(t)}\big).
\end{equation}
In TextGrad's terms, $(G_i, U)$ is the $i$-th module's step and the backward pass
over the pseudo-graph is the period's worth of rollouts: the loss node is the
evaluator's score, the intermediate nodes are the module outputs that conditioned
the candidate, and the leaf being optimized is $\theta_i$.

The RL-to-text correspondence:
\begin{center}\small
\begin{tabular}{@{}lll@{}}
\toprule
 & \textbf{RL} & \textbf{\sys} \\
\midrule
Policy parameters & network weights $\theta$ & module guidelines $\theta_i \in \mathcal{T}$ \\
Action            & $a_t \sim \pi_\theta(\cdot|s_t)$ & prefix fragment $m_i^{(t)}$ \\
Policy improvement& gradient step on $\theta$ & $U(\theta_i, G_i(\cdot))$ \\
\bottomrule
\end{tabular}
\end{center}
with the module's per-step behaviour given by
$m_i^{(t)} = \mathrm{LLM}\!\big(\theta_i^{(t)},\, s_t,\, M_4.\mathrm{memory}\big)$.

\subsection{Related systems}
\label{app:related}

\begin{table}[h]
\caption{\sys relative to the closest prior systems. The pattern is the same in
each row: prior work fixes a rule that our formalization turns into a component
optimized against $J$.}
\label{tab:related}
\begin{center}\small
\begin{tabular}{@{}p{2.6cm}p{5.0cm}p{5.0cm}@{}}
\toprule
\textbf{System} & \textbf{What it does} & \textbf{Difference in \sys} \\
\midrule
AlphaEvolve \citep{alphaevolve2025}   & Fixed prompt template + evolutionary search & Prefix is learnable \\
ShinkaEvolve \citep{shinkaevolve2025} & Bandit over LLMs + novelty sampling         & Full tool transfer, not only a bandit \\
EvoX \citep{evox2026}                 & Co-evolves solutions and search strategy    & Strategy improved by a critic against explicit $J$, not by selection \\
SMC-Evolve \citep{smcevolve2026}      & Principled sampling, fixed rules            & Rules are learned, not fixed \\
TextGrad \citep{textgrad2024}         & General-purpose text optimization           & Coupled to an RL objective in an evolutionary loop \\
RLHF / GRPO \citep{ouyang2022rlhf,shao2024grpo} & RL to train LLM parameters      & No parameter training; text-space optimization \\
\bottomrule
\end{tabular}
\end{center}
\end{table}

\subsection{The transfer}
\label{app:transfer}

\begin{table}[h]
\caption{The transfer. Each RL tool keeps its function but changes its form to
work without an empirical backup (column 4 is the honest column).}
\label{tab:transfer}
\begin{center}\scriptsize
\begin{tabular}{@{}p{1.6cm}p{2.0cm}p{1.5cm}p{4.2cm}@{}}
\toprule
\textbf{RL tool} & \textbf{Rule replaced} & \textbf{Module} & \textbf{What changed} \\
\midrule
Credit assignment & Success never attributed
  & $M_1$ & Textual attribution along genealogy, not numerical backup along chain \\
Value estimation & Must evaluate to rank
  & $M_2$ & Pre-evaluation \emph{ranking}; scalar $V$ estimated, not learned \\
Adaptive exploration & Fixed temperature
  & $M_3$ & Discrete explore/exploit decision + parent choice \\
Entropy regularization & Island diversity
  & $M_3$ (floor) & Explicit $\lambda H(\pi)$ in Eq.~\eqref{eq:reward} \\
Experience replay & Population \emph{is} memory
  & $M_4$ & Two tiers: durable abstract rules; per-direction path detail \\
Policy improvement & Hand-tuned hyperparams
  & Meta: TextGrad & Text-space improvement operator (\S\ref{sec:textgrad}) \\
\midrule
\emph{TD correction} & --- & --- & $\delta_t = r_t^{\text{true}} - \hat r_t$: formalized, not wired in \\
\bottomrule
\end{tabular}
\end{center}
\end{table}

\subsection{Limitations (full list)}
\label{app:limitations}

\begin{itemize}[leftmargin=1.4em,itemsep=1pt,topsep=2pt]
\item \textbf{Per-module attribution is unresolved.} The substrate ladder
(Table~\ref{tab:ladder}) reports end-to-end gain but no leave-one-out
decomposition of the four transferred modules; which of $M_1$--$M_4$ carries the
gain is open.
\item \textbf{$M_2$ is an estimator, not a value function.} It is not trained
against realized descendant fitness, so the $\gamma V_{M_2}$ term carries whatever
bias the text-space heuristic has, with no mechanism to correct it over the run.
\item \textbf{$V$'s ``orthogonality'' is partly self-inflicted.} $V$ correlates
negatively with fitness, which we read as carrying information fitness cannot see;
but $V$ is also what selects parents, and those descendants' gains are what define
$V$. The correlation is in part manufactured by the selection rule itself.
\item \textbf{$\gamma$ only discriminates among individuals with descendant
evidence.} $\texttt{lineage\_return}$ carries a horizon: programs born within the
last horizon iterations score $V{=}0$ and the ranking tie-breaks back to fitness,
so $\gamma$'s discriminative power applies to \emph{older} individuals.
\item \textbf{Unequal LLM budget.} At matched evaluations we spend
$2$--$4\times$ the baselines' calls; we report no LLM-call-matched arm.
\item \textbf{TextGrad has no convergence guarantee}, and its measured effect is
confounded by the redundant signal path identified in \S\ref{sec:textgrad}.
\item \textbf{$M_2$ bootstrapping.} The estimator calibrates itself with its own
estimates; we observe no divergence but offer no argument that it cannot occur.
\item \textbf{Critic and module share a backend model}, so self-evaluation blind
spots are possible; a cross-model critic is left to future work.
\item \textbf{Scope.} 3 seeds support qualitative conclusions only; 2 backend
models; main runs are single-worker.
\item \textbf{Baseline fairness.} Baseline systems are ported by us; our ports
are released so the conditions can be checked independently.
\end{itemize}

\subsection{Extensibility case study}
\label{app:extensibility}

The integration reported in \S\ref{sec:extensibility}(b) replaces $M_3$'s LLM
explore/exploit decision with the discounted-UCB bandit of
\citet{garivier2011ducb}. The arms are \textsc{exploit} and \textsc{explore};
on each update every arm's discounted pull count and reward sum decay by a
factor $g$ before the played arm is credited, and the decision is
$\arg\max_a\, s_a/n_a + c\sqrt{\log(N{+}1)/n_a}$. Cold start plays the
least-pulled arm until both have at least one discounted pull.

The diff is 236 lines in one new module file. The new class subclasses the
existing $M_3$ and overrides parent selection; it preserves the output contract
exactly (\texttt{result.selection} plus the \texttt{m3\_prefix} and
\texttt{m3\_decision} metadata of Appendix~\ref{app:modules}). Neither the
controller nor the module protocol is modified: the substrate contains zero
references to this mechanism, and it is selected by one configuration key.

Integrating it also \emph{deleted} 26 lines --- the hand-written budget and
stagnation decay heuristic that the discounting subsumes. That deletion is the
part we would point to: the claim of \S\ref{sec:extensibility} is not that new
mechanisms are cheap to add, but that adding a principled one can remove a
hand-tuned constant, which is the same claim the paper makes about the transfer
as a whole.

We note the limit of this case study honestly. The bandit is off by default and
is not used in any reported run (\S\ref{sec:discussion}), so what this
demonstrates is that the interface admits the mechanism at the stated cost ---
not that the mechanism helps.

\subsection{Per-task results, full detail}
\label{app:full-results}

Raw scores (mean over 3 seeds unless noted) for both backends. $\dagger$ = saturated (all systems within 1\%). $\diamond$ = fewer than 3 seeds completed.

\begin{center}\small
\begin{tabular}{@{}lrrrrc@{}}
\toprule
\textbf{Task} & OE & SE & \sys\,\textsf{Base} & \sys\,\textsf{Combo} & \textbf{win?} \\
\midrule
\multicolumn{6}{l}{\textit{Backend: DeepSeek-V4-Flash}} \\
\midrule
ahc008        & 0.6659 & 0.8511 & 0.5452 & 0.9041 & Combo \\
ahc011        & 0.4083 & 0.4635 & 0.4886 & 0.6515 & both \\
ahc015        & 0.8593 & 0.8087 & 0.9999$^\diamond$ & 0.8978$^\diamond$ & both \\
ahc016        & 0.6972 & 0.8602 & 0.9160 & 1.0188 & both \\
ahc024        & 0.6464 & 0.8327 & 0.8774 & 0.8636 & both \\
ahc025        & 0.6944 & 1.3799 & 2.616  & 3.0896 & both \\
ahc026        & 0.9671 & 0.9371 & 0.990  & 0.988  & both \\
ahc027        & 1.2421 & 1.5329 & 1.590  & 1.541  & both \\
ahc039        & 0.8748 & 0.9165 & 0.961  & 0.937  & both \\
ahc046        & 0.9729 & 0.9602 & 1.022  & 1.028  & both \\
circle$^\dagger$       & 0.9948 & 1.0001 & 0.9999 & 0.9994 & sat. \\
cloudcast$^\dagger$    & 0.0016 & 0.0016 & 0.0016 & 0.0016 & sat. \\
eplb          & 0.1519 & 0.1656 & 0.1643 & 0.1645 & --- \\
erdos$^\dagger$        & 0.9915 & 0.9997 & 0.9991 & 0.9966 & sat. \\
first\_ac$^\dagger$    & 0.9936 & 0.9959 & 0.9920 & 0.9942 & sat. \\
llm\_sql      & 0.7286 & 0.7322 & 0.7343 & 0.7436 & both \\
prism$^\dagger$        & 26.273 & 26.364 & 26.282 & 26.274 & sat. \\
second\_ac    & 0.9914 & 1.0099 & 1.0130 & 1.0369 & both \\
txn           & 4090.8 & 4392.7 & 4351   & 4419   & Combo \\
\midrule
\multicolumn{6}{l}{\textit{Backend: GLM-5.2}} \\
\midrule
ahc008        & 0.3700 & 0.5646 & 0.5965 & 0.6288 & both \\
ahc011        & 0.4213 & 0.3933 & 0.4483 & 0.4517 & both \\
ahc015        & 0.8472 & 0.8835 & 1.0355 & 0.9145 & both \\
ahc016        & 0.4476 & 0.4792 & 0.5985 & 0.6744 & both \\
ahc024        & 0.7120 & 0.8250 & 0.6968 & 0.8376 & Combo \\
ahc025        & 0.1803 & 0.2002 & 0.2734 & 0.2208 & both \\
ahc026        & 0.9742 & 0.9522 & 0.9745 & 0.9829 & both \\
ahc027        & 1.3306 & 1.1134 & 1.3183 & 1.2978 & --- \\
ahc039        & 0.5874 & 0.6232 & 0.7142 & 0.6408 & both \\
ahc046        & 0.9964 & 0.9621 & 1.0154 & 1.0136 & both \\
circle$^\dagger$       & 0.9939 & 0.9991 & 0.9969 & 0.9996 & sat. \\
cloudcast$^\dagger$    & 0.0016 & 0.0016 & 0.0016 & 0.0016 & sat. \\
eplb          & 0.1586 & 0.1678 & 0.1606 & 0.1611 & --- \\
erdos$^\dagger$        & 0.9996 & 0.9998 & 0.9995 & 0.9996 & sat. \\
first\_ac$^\dagger$    & 0.9931 & 0.9968 & 0.9973 & 0.9979 & sat. \\
llm\_sql      & 0.6797 & 0.7309 & 0.7407 & 0.7287 & Base \\
prism$^\dagger$        & 26.542 & 26.511 & 26.546 & 26.610 & sat. \\
second\_ac    & 1.0106 & 1.0274 & 1.0339 & 1.0185 & Base \\
txn           & 3990.6 & 4335.3 & 4374.7 & 4299.3 & Base \\
\bottomrule
\end{tabular}
\end{center}

\end{document}